\documentclass[
a4paper,orivec]{llncs}

\usepackage{amssymb}
\usepackage{graphicx}

\usepackage{url}
\urldef{\mailsa}\path|jose.m.pena@liu.se|

\usepackage{graphicx,amssymb,datetime,float,MnSymbol,currfile,multirow}
\usepackage{tikz}
\usetikzlibrary{arrows}

\def\ci{\!\perp\!}

\def\ra{\rightarrow}
\def\la{\leftarrow}

\newcommand{\comments}[1]{}

\begin{document}

\mainmatter  

\title{Factorization, Inference and Parameter Learning in Discrete AMP Chain Graphs}

\titlerunning{Factorization, Inference and Parameter Learning in Discrete AMP Chain Graphs}

%
%
\author{Jose M. Pe\~{n}a}
\authorrunning{Factorization, Inference and Parameter Learning in Discrete AMP Chain Graphs}

\institute{ADIT, IDA, Link\"oping University, SE-58183 Link\"{o}ping, Sweden\\
\mailsa}

%
%

\toctitle{Lecture Notes in Computer Science}
\tocauthor{Authors' Instructions}
\maketitle

\begin{abstract}
We address some computational issues that may hinder the use of AMP chain graphs in practice. Specifically, we show how a discrete probability distribution that satisfies all the independencies represented by an AMP chain graph factorizes according to it. We show how this factorization makes it possible to perform inference and parameter learning efficiently, by adapting existing algorithms for Markov and Bayesian networks. Finally, we turn our attention to another issue that may hinder the use of AMP CGs, namely the lack of an intuitive interpretation of their edges. We provide one such interpretation.
\end{abstract}

\section{Introduction}

Chain graphs (CGs) are graphs with possibly directed and undirected edges, and no semidirected cycle. They have been extensively studied as a formalism to represent independence models, because they can model symmetric and asymmetric relationships between random variables. There are three different interpretations of CGs as independence models: The Lauritzen-Wermuth-Frydenberg (LWF) interpretation \cite{Lauritzen1996}, the multivariate regression (MVR) interpretation \cite{CoxandWermuth1996}, and the Andersson-Madigan-Perlman (AMP) interpretation \cite{Anderssonetal.2001}. No interpretation subsumes another \cite{Anderssonetal.2001,SonntagandPenna2015}.

In this paper, we focus on AMP CGs. Despite being much more expressive than Markov and Bayesian networks \cite{Sonntag2014}, AMP CGs have not enjoyed much success in the literature or in practice. We believe this is due to mainly two reasons. First, it is not known how to perform inference and parameter learning for AMP CGs efficiently, because it is now known how to factorize a probability distribution that satisfies all the independencies represented by an AMP CG. Compare this situation to that of LWF CGs, where such a factorization exists \cite[Theorem 4.1]{Frydenberg1990} and thus inference can be performed efficiently \cite[Section 6.5]{Cowelletal.1999}. Second, AMP CGs do not appeal to intuition: Whereas the directed edges in a Bayesian network may be interpreted as causal relationships and the undirected edges in a Markov network as correlation relationships, it is not clear how to combine these two interpretations to produce an intuitive interpretation of the edges in an AMP CG.

In this paper, we address the two problems mentioned above. First, we introduce a factorization for AMP CGs and show how it makes it possible to perform inference and parameter learning efficiently, by adapting existing algorithms for Markov and Bayesian networks. Second, we propose an intuitive interpretation of the edges in an AMP CG. We start with some notation and definitions.

\section{Preliminaries}

Unless otherwise stated, all the graphs and probability distributions in this paper are defined over a finite set of discrete random variables $V$. We use uppercase letters to denote random variables and lowercase letters to denote their states. The elements of $V$ are not distinguished from singletons. If a graph $G$ contains an undirected or directed edge between two nodes $V_{1}$ and $V_{2}$, then we write that $V_{1} - V_{2}$ or $V_{1} \ra V_{2}$ is in $G$. The parents of a set of nodes $X$ of $G$ is the set $Pa_G(X) = \{V_1 | V_1 \ra V_2$ is in $G$, $V_1 \notin X$ and $V_2 \in X \}$. The adjacents of $X$ is the set $Ad_G(X) = \{V_1 | V_1 \la V_2$, $V_1 \ra V_2$ or $V_1 - V_2$ is in $G$, $V_1 \notin X$ and $V_2 \in X \}$. A route between a node $V_{1}$ and a node $V_{n}$ in $G$ is a sequence of (not necessarily distinct) nodes $V_{1}, \ldots, V_{n}$ st $V_i \in Ad_G(V_{i+1})$ for all $1 \leq i < n$. If the nodes in the route are all distinct, then the route is called a path. A route is called descending if $V_i \ra V_{i+1}$ or $V_i - V_{i+1}$ is in $G$ for all $1 \leq i < n$. A route is called strictly descending if $V_i \ra V_{i+1}$ is in $G$ for all $1 \leq i < n$. The descendants of a set of nodes $X$ of $G$ is the set $De_G(X) = \{V_n |$ there is a descending route from $V_1$ to $V_n$ in $G$, $V_1 \in X$ and $V_n \notin X \}$. The non-descendants of $X$ is the set $Nd_G(X)=V \setminus X \setminus De_G(X)$. The strict ascendants of $X$ is the set $Sa_G(X) = \{V_1 |$ there is a strictly descending route from $V_1$ to $V_n$ in $G$, $V_1 \notin X$ and $V_n \in X \}$. A route $V_{1}, \ldots, V_{n}$ in $G$ is called a cycle if $V_n=V_1$. Moreover, it is called a semidirected cycle if $V_n=V_1$, $V_1 \ra V_2$ is in $G$ and $V_i \ra V_{i+1}$ or $V_i - V_{i+1}$ is in $G$ for all $1 < i < n$. An AMP chain graph (AMP CG) is a graph whose every edge is directed or undirected st it has no semidirected cycles. An AMP CG with only directed edges is called a directed and acyclic graph (DAG), whereas an AMP CG with only undirected edges is called an undirected graph (UG). A set of nodes of an AMP CG $G$ is connected if there exists a route in the CG between every pair of nodes in the set st all the edges in the route are undirected. A connectivity component of $G$ is a maximal (wrt set inclusion) connected set of nodes. The connectivity components of $G$ are denoted as $Cc(G)$, whereas $Cc_G(X)$ denotes the connectivity component to which the node $X$ belongs. A set of nodes of $G$ is complete if there exists an undirected edge between every pair of nodes in the set. The complete sets of nodes of $G$ are denoted as $Cs(G)$. A clique of $G$ is a maximal (wrt set inclusion) complete set of nodes. The cliques of $G$ are denoted as $Cl(G)$. The subgraph of $G$ induced by a set of its nodes $X$, denoted as $G_X$, is the graph over $X$ that has all and only the edges in $G$ whose both ends are in $X$.

We now recall the semantics of AMP CGs. A node $B$ in a path $\rho$ in an AMP CG $G$ is called a triplex node in $\rho$ if $A \ra B \la C$, $A \ra B - C$, or $A - B \la C$ is a subpath of $\rho$. Moreover, $\rho$ is said to be $Z$-open with $Z \subseteq V$ when
\begin{itemize}
\item every triplex node in $\rho$ is in $Z \cup Sa_G(Z)$, and

\item every non-triplex node $B$ in $\rho$ is outside $Z$, unless $A - B - C$ is a subpath of $\rho$ and $Pa_G(B) \setminus Z \neq \emptyset$.
\end{itemize}

Let $X$, $Y$ and $Z$ denote three disjoint subsets of $V$. When there is no $Z$-open path in an AMP CG $G$ between a node in $X$ and a node in $Y$, we say that $X$ is separated from $Y$ given $Z$ in $G$ and denote it as $X \ci_G Y | Z$. The independence model represented by $G$ is the set of separations $X \ci_G Y | Z$. The independence model represented by $G$ under marginalization of some nodes $L \subseteq V$ is the set of separations $X \ci_G Y | Z$ with $X, Y, Z \subseteq V \setminus L$. Finally, we denote by $X \ci_p Y | Z$ that $X$ is independent of $Y$ given $Z$ in a probability distribution $p$. We say that $p$ is Markovian wrt an AMP CG $G$ when, for all $X$, $Y$ and $Z$ disjoint subsets of $V$, if $X \ci_G Y | Z$ then $X \ci_p Y | Z$.

\section{Factorization}\label{sec:factorization}

A probability distribution $p$ is Markovian wrt an AMP CG $G$ iff the following three properties hold for all $C \in Cc(G)$ \cite[Theorem 2]{Anderssonetal.2001}:
\begin{itemize}
\item C1: $C \ci_p Nd_G(C) \setminus Cc_G(Pa_G(C)) | Cc_G(Pa_G(C))$.

\item C2: $p(C | Cc_G(Pa_G(C)))$ is Markovian wrt $G_C$.

\item C3$^*$: For all $D \subseteq C$, $D \ci_p Cc_G(Pa_G(C)) \setminus Pa_G(D) | Pa_G(D)$.
\end{itemize}

Then, C1 implies that $p$ factorizes as
\[
p = \prod_{C \in Cc(G)} p(C | Cc_G(Pa_G(C))).
\]
The authors of \cite[p. 50]{Anderssonetal.2001} note that if $p$ were strictly positive and $G$ were a LWF CG, then each conditional distribution above would factorize further into a product of potentials over certain subsets of the nodes in $C \cup Pa_G(C)$, as shown in \cite[Theorem 4.1]{Frydenberg1990}. However, the authors state that no such further factorization appears to hold in general if $G$ is an AMP CG. We show that this is not true if $p$ is strictly positive. Specifically, C2 together with \cite[Theorems 3.7 and 3.9]{Lauritzen1996} imply that
\[
p(C | Cc_G(Pa_G(C))) = \prod_{K \in Cs(G_C)} \varphi(K, Cc_G(Pa_G(C))).
\]
However, one can show that $\varphi(K, Cc_G(Pa_G(C)))$ is actually a function of $K \cup Pa_G(K)$, i.e. $\varphi(K, Cc_G(Pa_G(C))) = \varphi(K, Pa_G(K))$. It suffices to recall from the proof of \cite[Theorem 3.9]{Lauritzen1996} how $\varphi(K, Cc_G(Pa_G(C)))$ can be obtained from $p(C | Cc_G(Pa_G(C)))$, a method also known as canonical parameterization \cite[Section 4.4.2.1]{KollerandFriedman2009}. Specifically, let $\phi(K, Cc_G(Pa_G(C))) = \log \varphi(K, Cc_G(Pa_G(C)))$. Choose a fixed but arbitrary state $k_*$ of $K$. Then,
\[
\phi(k, Cc_G(Pa_G(C))) = \sum_{q \subseteq k} (-1)^{|k \setminus q|} \log p(q, \overline{q}_* | Cc_G(Pa_G(C)))
\]
where $\overline{q}_*$ denotes the elements of $k_*$ corresponding to the elements of $K \setminus Q$. Now, note that $p(q, \overline{q}_* | Cc_G(Pa_G(C))) = p(q, \overline{q}_* | Pa_G(K))$ by C3$^*$, because $Q \subseteq K$. Then, $\varphi(K, Cc_G(Pa_G(C)))$ is actually a function of $K \cup Pa_G(K)$. 

Putting together the results above, we have that $p$ factorizes as
\begin{equation}\label{equ:factorization}
p = \prod_{C \in Cc(G)} \prod_{K \in Cs(G_C)} \varphi(K, Pa_G(K)) = \prod_{C \in Cc(G)} \prod_{K \in Cl(G_C)} \psi(K, Pa_G(K)).
\end{equation}

Note that the well-known factorizations induced by DAGs and UGs (see \cite[Sections 3.2.1 and 3.2.2]{Lauritzen1996}) are special cases of Equation \ref{equ:factorization}.

\section{Parameter Learning}\label{sec:parameter}

The factorization in Equation \ref{equ:factorization} enables us to perform parameter learning for AMP CGs efficiently by deploying the iterative proportional fitting procedure (IPFP) \cite[Section 19.5.7]{Murphy2012}, which returns the maximum likelihood estimates of the entries of the potentials for some given data. Specifically, we first simplify further the factorization by multiplying its potentials until no potential domain is included in another potential domain. Let $Q_1, \ldots, Q_n$ denote the potential domains in the resulting factorization. Note that each domain $Q_i$ is of the form $K \cup Pa_G(K)$ with $K \in Cl(G_C)$ and $C \in Cc(G)$. Then, we run the IPFP per se:
\begin{center}
\begin{tabular}{rl}
1 & For each potential $\psi(Q_i)$\\
2 & \hspace{0.25cm}  Set $\psi^0(Q_i) = 1$\\
3 & Repeat until convergence\\
4 & \hspace{0.25cm}  For each potential $\psi^t(Q_i)$\\
5 & \hspace{0.5cm} Set $\psi^{t+1}(Q_i) = \psi^t(Q_i) \frac{p_e(Q_i)}{p^t(Q_i)}$\\
\end{tabular}
\end{center}
where $p^t = \prod_{i=1}^n \psi^t(Q_i)$, and $p_e$ is the empirical probability distribution over $V$ obtained from the given data.

\section{Inference}\label{sec:inference}

The factorization in Equation \ref{equ:factorization} also enables us to perform inference in AMP CGs efficiently by deploying the algorithm for inference in DAGs developed by \cite{LauritzenandSpiegelhalter1988}, and upon which most other inference algorithms build. Specifically, we start by transforming $G$ into its moral graph $G^m$ by running the procedure below. This procedure differs from the one in \cite{LauritzenandSpiegelhalter1988}, because $G$ is an AMP CG and not a DAG. In any case, the moralization procedure in \cite{LauritzenandSpiegelhalter1988} is a special case of the procedure below.
\begin{center}
\begin{tabular}{rl}
1 & Set $G^m=G$\\
2 & For each connectivity component $C \in Cc(G)$\\
3 & \hspace{0.25cm} For each clique $K \in Cl(G_C)$\\
4 & \hspace{0.5cm} Add the edge $X \ra Y$ to $G^m$ for all $X \in Pa_G(K)$ and $Y \in K$\\
5 & \hspace{0.5cm} Add the edge $X - Y$ to $G^m$ for all $X, Y \in Pa_G(K)$ st $X \neq Y$\\
6 & Replace all the directed edges in $G^m$ with undirected edges\\
\end{tabular}
\end{center}

The reason of why $G^m$ has the edges it has will become clear later. We continue by transforming $G^m$ into a triangulated graph $G^t$, and sorting its cliques to satisfy the so-called running intersection property. The procedure below accomplishes these two objectives. An UG is triangulated when every cycle in it contains a chord, i.e. an edge between two non-consecutive nodes in the cycle. The cliques of a triangulated graph can be ordered as $Q_1, \ldots, Q_n$ so that for all $1 < j \leq n$, $Q_j \cap (Q_1 \cup \ldots \cup Q_{j-1}) \subseteq Q_i$ for some $1 \leq i < j$. This is known as the running intersection property (RIP).
\begin{center}
\begin{tabular}{rl}
1 & Set $G^t=G^m$\\
2 & Repeat until all the nodes in $G^t$ are marked\\
3 & \hspace{0.25cm} Select an unmarked node in $G^t$ with the largest number of marked\\
& \hspace{0.25cm} neighbours\\
4 & \hspace{0.25cm} Mark the node and make its marked neighbours form a complete set\\
& \hspace{0.25cm} in $G^t$ by adding undirected edges\\
5 & \hspace{0.25cm} Save the node plus its marked neighbours as a candidate clique\\
6 & Remove every candidate clique that is included in another\\
7 & Label every clique with the last iteration that marked one of its nodes\\
8 & Sort the cliques in ascending order of their labels\\
\end{tabular}
\end{center}

Finally, let $Q_1, \ldots, Q_n$ denote the ordering of the cliques of $G^t$ returned by the procedure above. Let $S_j = Q_j \cap (Q_1 \cup \ldots \cup Q_{j-1})$ and $R_j = Q_j \setminus S_j$. Note that for every $K \in Cl(G_C)$ with $C \in Cc(G)$, there is some $Q_i$ st $K \cup Pa_G(K) \subseteq Q_i$, because the moralization procedure above made $K \cup Pa_G(K)$ a complete set in $G^m$ and thus in $G^t$. Then,
\begin{equation}\label{equ:1}
p(V) = \prod_{C \in Cc(G)} \prod_{K \in Cl(G_C)} \psi(K, Pa_G(K)) = \prod_{i=1}^n \phi(Q_i)
\end{equation}
and thus 
\[
p(V)=f([ Q_1 \cup \ldots \cup Q_{n-1} ] \setminus S_n, S_n) g(S_n, R_n)
\]
and thus
\[
R_n \ci_p [ Q_1 \cup \ldots \cup Q_{n-1} ] \setminus S_n | S_n
\]
by \cite[p. 29]{Lauritzen1996}, and thus
\begin{equation}\label{equ:2}
p(V) = p(Q_1 \cup \ldots  \cup Q_{n-1}) p(R_n | Q_1 \cup \ldots \cup Q_{n-1}) = p(Q_1 \cup \ldots \cup Q_{n-1}) p(R_n | S_n).
\end{equation}
Note also that
\begin{equation}\label{equ:3}
p(Q_1 \cup \ldots \cup Q_{n-1}) = \sum_{r_n} p(Q_1 \cup \ldots \cup Q_{n-1}, r_n) = [ \prod_{i=1}^{n-1} \phi(Q_i) ] \sum_{r_n} \phi(S_n, r_n).
\end{equation}
Then, Equations \ref{equ:1}-\ref{equ:3} imply that
\[
p(R_n | S_n) = \phi(Q_n) / \sum_{r_n} \phi(S_n, r_n).
\]
Note that $S_n \subseteq Q_j$ for some $1 \leq j<n$ by the RIP. Then, we replace $\phi(Q_j)$ with $\phi(Q_j) \sum_{r_n} \phi(S_n, r_n)$, after which Equation \ref{equ:3} implies that
\[
p(Q_1 \cup \ldots \cup Q_{n-1}) = \prod_{i=1}^{n-1} \phi(Q_i).
\]
We repeat the steps above for $p(Q_1 \cup \ldots \cup Q_{n-1})$ and so we obtain $p(R_i | S_i)$ for all $1 \leq i \leq n$. Now, note that $S_1 = \emptyset$ and, thus, $p(Q_1)=p(R_1 | S_1)$. Moreover, since $S_2 \subseteq Q_1$ by the RIP, then
\[
p(S_2) = \sum_{q_1 \setminus s_2} p(S_2, q_1 \setminus s_2)
\]
and thus
\[
p(Q_2)=p(R_2 | S_2) p(S_2).
\]
We repeat the steps above for $Q_3, \ldots, Q_n$ and so we obtain $p(Q_i)$ for all $1 \leq i \leq n$. To obtain $p(Q_i | o)$ where $o$ denotes some observations or evidence, we first remove all the entries of $\phi(Q_j)$ that are inconsistent with $o$ for all $1 \leq j \leq n$, then we repeat the steps above to get $p(Q_i, o)$ and, finally, we normalize by $p(o)=\sum_{q_i} p(q_i, o)$. To obtain $p(X | o)$ where $X \nsubseteq Q_i$ for all $1 \leq i \leq n$, we compute $p(x, o)$ for all $x$ as if \{x, o\} were the observations and, then, we normalize by $p(o)=\sum_x p(x, o)$.

\section{Error AMP CGs}

So far in this article, we have shown how an AMP CG factorizes a probability distribution, and how this helps in performing parameter learning and inference efficiently. We believe that our findings solve some computational issues that have hindered the use of AMP CGs in practice. In this section, we turn our attention to another issue that may have also hindered the use of AMP CGs, namely the lack of an intuitive interpretation of their edges. Whereas the directed edges in a DAG may be interpreted as causal relationships and the undirected edges in an UG as correlation relationships, it is not clear how to combine these two interpretations to produce an intuitive interpretation of the edges in an AMP CG. We propose here a way to do it by adapting to discrete AMP CGs the interpretation for Gaussian AMP CGs presented in \cite[Section 5]{Anderssonetal.2001} and further studied in \cite[Section 3]{Penna2014}. Specifically, we propose to interpret the directed edges in an AMP CG as causal relationships. In other words, the parents of a node represent its causal mechanism. We propose to assume that this mechanism is deterministic but it may sometimes work erroneously. We propose to interpret the undirected edges in the AMP CG as the correlation structure of the errors of the causal mechanisms of the different nodes. To show the validity of this interpretation, we will first modify the AMP CG by adding a deterministic node for each original node to represent explicitly the occurrence or not of an error in its causal mechanism and, then, we will show that the original and the modified AMP CGs are equivalent in some sense. We call the modified CG an error AMP (EAMP) CG. Since an EAMP CG is an AMP CG with deterministic nodes, we discuss these first.

\subsection{AMP CGs with Deterministic Nodes}\label{sec:deterministic}

We say that a node $X$ of an AMP CG is determined by some $Z \subseteq V$ when $X \in Z$ or $X$ is a function of $Z$ in each probability distribution that is Markovian wrt the CG. In that case, we also say that $X$ is a deterministic node. We use $D(Z)$ to denote all the nodes that are determined by $Z$. From the point of view of the separations in an AMP CG, that a node outside the conditioning set of a separation is determined by it, has the same effect as if the node were actually in the conditioning set. We extend accordingly the definition of separation for AMP CGs to the case where deterministic nodes may exist. Given an AMP CG $G$, a path $\rho$ in $G$ is said to be $Z$-open when
\begin{itemize}
\item every triplex node in $\rho$ is in $D(Z) \cup Sa_G(D(Z))$, and

\item no non-triplex node $B$ in $\rho$ is in $D(Z)$, unless $A - B - C$ is a subpath of $\rho$ and $Pa_G(B) \setminus D(Z) \neq \emptyset$.
\end{itemize}

\begin{figure}[t]
\begin{center}
\begin{tabular}{|cc|cc|}
\hline
\begin{tikzpicture}[inner sep=1mm]
\node at (0,0) (A) {$A$};
\node at (1,0) (B) {$B$};
\node at (0,-1) (C) {$C$};
\node at (1,-1) (D) {$D$};
\node at (0,-3) (E) {$F$};
\node at (1,-3) (F) {$I$};
\path[->] (A) edge (B);
\path[->] (A) edge (C);
\path[->] (A) edge (D);
\path[->] (B) edge (D);
\path[-] (C) edge (D);
\path[-] (C) edge (E);
\path[-] (D) edge (F);
\path[-] (E) edge (F);
\end{tikzpicture}
&&&
\begin{tikzpicture}[inner sep=1mm]
\node at (0,0) (A) {$A$};
\node at (1,0) (B) {$B$};
\node at (0,-1) (C) {$C$};
\node at (1,-1) (D) {$D$};
\node at (0,-3) (E) {$F$};
\node at (1,-3) (F) {$I$};
\node at (-1,0) (EA) {$E_A$};
\node at (2,0) (EB) {$E_B$};
\node at (-1,-1) (EC) {$E_C$};
\node at (2,-1) (ED) {$E_D$};
\node at (-1,-3) (EE) {$E_F$};
\node at (2,-3) (EF) {$E_I$};
\path[->] (EA) edge (A);
\path[->] (EB) edge (B);
\path[->] (EC) edge (C);
\path[->] (ED) edge (D);
\path[->] (EE) edge (E);
\path[->] (EF) edge (F);
\path[->] (A) edge (B);
\path[->] (A) edge (C);
\path[->] (A) edge (D);
\path[->] (B) edge (D);
\path[-] (EC) edge [bend right] (ED);
\path[-] (EC) edge (EE);
\path[-] (ED) edge (EF);
\path[-] (EE) edge [bend left] (EF);
\end{tikzpicture}\\
\hline
\end{tabular}
\end{center}\caption{An AMP CG and its corresponding EAMP CG.}\label{fig:example}
\end{figure}
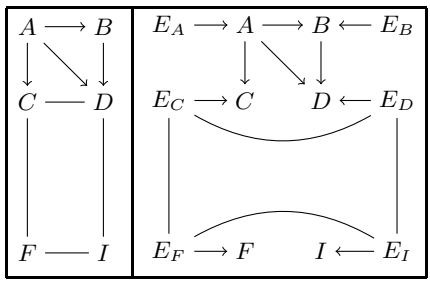

\subsection{EAMP CGs}

The EAMP CG $H$ corresponding to an AMP CG $G$ is an AMP CG over $V \cup E$, where $E$ denotes the error nodes. Specifically, there is an error node $E_X \in E$ for every node $X \in V$, and it represents whether an error in the causal mechanism of $X$ occurs or not. We set $Pa_H(X) = Pa_G(X) \cup E_X$ to represent that $E_X$ is part of the causal mechanism of $X$ in $H$. This causal mechanism works as follows: If $E_X=0$ (i.e. no error) then $pa_G(X)$ determines the state of $X$ to be the distinguished state $x_*^{pa_G(X)}$, else $X$ may take any state but the distinguished one. The undirected edges in $H$ are all between error nodes, and they represent the correlation structure of the error nodes. Specifically, the undirected edge $E_X - E_Y$ is in $H$ iff the undirected edge $X - Y$ is in $G$. Note that the error nodes are never observed, i.e. they are latent. The procedure below formalizes the transformation just described. See Figure \ref{fig:example} for an example. 
\begin{center}
\begin{tabular}{rl}
1 & Set $H=G$\\
2 & For each node $X \in V$\\
3 & \hspace{0.25cm} Add the node $E_X$ and the edge $E_X \ra X$ to $H$\\
4 & Replace every edge $X - Y$ in $H$ st $X, Y \in V$ with an edge $E_X - E_Y$\\
\end{tabular}
\end{center}
Now, consider a probability distribution $p(V, E)$ that is Markovian wrt the EAMP CG $H$. Then,
\begin{equation}\label{equ:factorization2}
p(V, E) = p(V | E) p(E) = [ \prod_{X \in V} p(X | Pa_G(X), E_X) ] p(E)
\end{equation}
by C1 and C3$^*$. Moreover, in order for the causal mechanism of $X$ in $H$ to match the description above, we restrict $p(X | Pa_G(X), E_X)$ to be of the following form: 
\begin{equation}\label{equ:constraint}
p(X | pa_G(X), E_X) = \left\{
\begin{array}{l l}
1 & \text{if $E_X = 0$ and $X = x_*^{pa_G(X)}$}\\
0 & \text{if $E_X = 0$ and $X \neq x_*^{pa_G(X)}$}\\
q(X | pa_G(X)) & \text{if $E_X = 1$}\\
\end{array} \right.
\end{equation}
where $q(X | pa_G(X))$ is an arbitrary conditional probability distribution with the only constraints that $q(X | pa_G(X)) = 0$ if $X = x_*^{pa_G(X)}$, and $q(X | pa_G(X)) > 0$ otherwise. The first constraint follows from the description above of the causal mechanism of $X$ in $H$, whereas the second is necessary for $p(V)$ being strictly positive. Note that $E_X$ is determined by $Pa_G(X) \cup X$. Specifically, if $X = x_*^{pa_G(X)}$ then $E_X = 0$, else $E_X = 1$. Then, $E$ is determined by $V$. Hereinafter, when we say that a probability distribution is Markovian wrt an EAMP CG, it should be understood that it also satisfies the constraint in Equation \ref{equ:constraint}.

We assume that $p(E)$ is strictly positive, as a way to ensure that $p(V)$ is strictly positive. This together with the fact that $p(E)$ is Markovian wrt $H_E$, which follows from $p(V, E)$ being Markovian wrt $H$, implies that $p(E)$ factorizes as shown in Equation \ref{equ:factorization} and, thus, Equation \ref{equ:factorization2} becomes
\begin{equation}\label{equ:factorization3}
p(V, E) = [ \prod_{X \in V} p(X | Pa_G(X), E_X) ] [ \prod_{E_C \in Cc(H_{E})} \prod_{E_K \in Cl(H_{E_C})} \phi(E_K) ].
\end{equation}
Thus, it is clear that the EAMP CG $H$ can be interpreted as we wanted: Each node is controlled by the causal mechanism specified in the AMP CG $G$, the mechanism is deterministic if no error occurs and it is random otherwise, and the errors of the different mechanisms obey the correlation structure specified in $G$. To see the last point, note that $E_C \in Cc(H_{E})$ iff $C \in Cc(G)$, and $E_K \in Cl(H_{E_C})$ iff $K \in Cl(G_C)$. Thus, $H$ somehow keeps the structural information in $G$. To make this claim more specific, note that the independence model represented by $G$ coincides with that represented by $H$ under marginalization of the error nodes which, recall from above, are latent \cite[Theorem 1]{Penna2014}.\footnote{Unlike in this work, $V$ is a Gaussian random variable in \cite{Penna2014}. However, that is irrelevant in the proof of \cite[Theorem 1]{Penna2014}. The proof builds upon the following two properties which, as we show, also hold for the framework in this work:
\begin{itemize}
\item A node $E_X \in E$ is determined by some $Z \subseteq V$ iff $Pa_G(X) \cup X \subseteq Z$. The if part follows from the fact shown above that $E_X$ is determined by $Pa_G(X) \cup X$. To see the only if part, assume to the contrary that $Z$ determines $E_X$ but $Pa_G(X) \cup X \nsubseteq Z$. Then, $X \notin Z$ or there is some $Y \in Pa_G(X) \setminus Z$. If $X \notin Z$, then let $H'$ be the EAMP CG $H'$ over $V \cup E$ whose only edge is $E_X \ra X$, and let $p'$ be a probability distribution that is Markovian wrt $H'$. Note that $E_X$ is a function of just $X$ in $p'$. If $X \in Z$, then let $H'$ have the edges $E_X \ra X \la Y \la E_Y$, and let $p'$ be Markovian wrt $H'$ st $x_*^{y_0} \neq x_*^{y_1}$. Note that $E_X$ is a function of just $X \cup Y$ in $p'$. Note also that in either case $p'$ is Markovian wrt $H$, because $H'$ is a subgraph of $H$. Note also that in neither case $E_X$ is a function of $Z$ in $p'$. This contradicts that $Z$ determines $E_X$.

\item A node $X \in V$ is determined by some $Z \subseteq V$ iff $X \in Z$. The if part is trivial. To see the only if part, note that $X$ is determined by $Z$ only if $X \in Z$ or $E_X$ is determined by $Z$. However, $E_X$ is determined by $Z$ only if $X \in Z$ by the previous property.
\end{itemize}} Recall that the independence model represented by $H$ can be read off as shown in Section \ref{sec:deterministic}. Note that that the independence model represented by $G$ coincides with that represented by $H$ under marginalization of the error nodes implies that the probability distribution resulting from marginalizing $E$ out of a distribution $p(V, E)$ that is Markovian wrt to $H$ is Markovian wrt $G$ and, thus, it factorizes as shown in Equation \ref{equ:factorization}. Specifically, recall that $E$ is determined by $V$ and, thus, $p(V, E)$ is actually a function of $V$. Then, it suffices to set each potential $\psi(K, Pa_G(K))$ in Equation \ref{equ:factorization} equal to the following product of the terms in Equation \ref{equ:factorization3}:
\[
\psi(K, Pa_G(K)) = [ \prod_{X \in K} p(X | Pa_G(X), E_X) ] \phi(E_K)
\]
bearing in mind that if $X$ belongs to several cliques $K$, then $p(X | Pa_G(X), E_X)$ is assigned to only one (any) of the potentials $\psi(K, Pa_G(K))$. For instance, the following is a valid assignment for the AMP and EAMP CGs in Figure \ref{fig:example}:
\[
\begin{array}{l l l | l l l}
\psi(A) &=& p(A | E_A) \phi(E_A) & \psi(C, F, A) &=& p(F | E_F) \phi(E_C, E_F)\\
\psi(B, A) &=& p(B | A, E_B) \phi(E_B) & \psi(D, I, A, B) &=& p(D | A, B, E_D) \phi(E_D, E_I)\\
\psi(C, D, A, B) &=& p(C | A, E_C) \phi(E_C, E_D) & \psi(F, I) &=& p(I | E_I) \phi(E_F, E_I)\\
\end{array}
\]

Unfortunately, the opposite of the last result above does not hold. That is, not every probability distribution that factorizes according to an AMP CG coincides with the marginal of a distribution that is Markovian wrt the corresponding EAMP CG. To see it, let $G$ be the AMP CG $A \ra B - C$. Let $H$ be the EAMP CG corresponding to $G$, i.e. $E_A \ra A \ra B \la E_B - E_C \ra C$. Consider a probability distribution $p(A, B, C, E_A, E_B, E_C)$ that is Markovian wrt $H$. Since as shown above $\{E_A, E_B, E_C\}$ is determined by $\{A, B, C\}$, Equation \ref{equ:factorization3} implies that
\begin{equation}\label{equ:example}
\frac{p(a_0, b_*^{a_0}, C)}{p(a_1, b_*^{a_1}, C)} = \frac{p(a_0 | E_A) p(b_*^{a_0} | a_0, E_B) p(C | E_C) \phi(E_A) \phi(E_B, E_C)}{p(a_1 | E_A) p(b_*^{a_1} | a_1, E_B) p(C | E_C) \phi(E_A) \phi(E_B, E_C)} = \frac{p(a_0 | E_A) \phi(E_A)}{p(a_1 | E_A) \phi(E_A)}
\end{equation}
because both $\{ a_0, b_*^{a_0} \}$ and $\{ a_1, b_*^{a_1} \}$ determine that $E_B=0$, which implies that $p(b_*^{a_0} | a_0, E_B) = p(b_*^{a_1} | a_1, E_B) = 1$. Now, consider a probability distribution $p'(A, B, C)$ that factorizes according to $G$. Then, Equation \ref{equ:factorization} implies that
\begin{equation}\label{equ:example2}
\frac{p'(a_0, b_*^{a_0}, C)}{p'(a_1, b_*^{a_1}, C)} = \frac{\psi(a_0) \psi(a_0, b_*^{a_0}, C)}{\psi(a_1) \psi(a_1, b_*^{a_1}, C)}.
\end{equation}
Note that the ratio in Equation \ref{equ:example2} is a function of $C$ whereas the ratio in Equation \ref{equ:example} is not. Therefore, $p(A, B, C) \neq p'(A, B, C)$ in general.

Finally, note that every node $X \in V$ in an EAMP CG $H$ forms a connectivity component on its own. Therefore, the factorization in Equation \ref{equ:factorization3} is actually of the same form as the factorization in Equation \ref{equ:factorization}. This comes as no surprise because, after all, $H$ is an AMP CG over $V \cup E$.

\section{Discussion}

We have addressed some issues that may hinder the use of AMP CGs in practice. We hope that the results reported in this paper help others to deploy AMP CGs in practical applications. Specifically, we have shown how a discrete probability distribution that is Markovian wrt an AMP CG factorizes according to it. We have also shown how this factorization makes it possible to perform inference and parameter learning efficiently. Finally, we have provided an intuitive interpretation of AMP CGs that sheds some light on what the different edges may mean. Unfortunately, the interpretation provided is not perfect, i.e. not every probability distribution that factorizes according to an AMP CG coincides with the marginal of a distribution that is Markovian wrt the corresponding EAMP CG. We are working to solve this problem. We are also working on proving the opposite of the result in Section \ref{sec:factorization}, i.e. proving that every probability distribution that factorizes according to an AMP CG is Markovian wrt it.

\subsubsection*{Acknowledgments.}

This work is funded by the Swedish Research Council (ref. 2010-4808), and by a so-called career contract at Link\"oping University.

\end{document}


\title[]{Factorization, Inference and Parameter Learning in Discrete AMP Chain Graphs: Addendum}

\author[]{Jose M. Pe\~{n}a\\
ADIT, IDA, Link\"oping University, SE-58183 Link\"{o}ping, Sweden\\
jose.m.pena@liu.se}

\date{\currfilename, \currenttime, \ddmmyydate{\today}}

\maketitle

This note extends the original manuscript with new results, and corrects some errors.

\section{Factorization}

A probability distribution $p$ is Markovian wrt an AMP CG $G$ iff the following three properties hold for all $C \in Cc(G)$ \cite[Theorem 2]{Anderssonetal.2001}:
\begin{itemize}
\item C1: $C \ci_p Nd_G(C) \setminus Cc_G(Pa_G(C)) | Cc_G(Pa_G(C))$.

\item C2: $p(C | Cc_G(Pa_G(C)))$ is Markovian wrt $G_C$.

\item C3$^*$: For all $D \subseteq C$, $D \ci_p Cc_G(Pa_G(C)) \setminus Pa_G(D) | Pa_G(D)$.
\end{itemize}

\begin{lemma}
C1, C2 and C3$^*$ hold iff the following two properties hold:
\begin{itemize}
\item C1$^*$: For all $D \subseteq C$, $D \ci_p Nd_G(D) \setminus Pa_G(D) | Pa_G(D)$.

\item C2$^*$: $p(C | Pa_G(C))$ is Markovian wrt $G_C$.
\end{itemize}
\end{lemma}

\begin{proof}
First, C1$^*$ implies C3$^*$ by decomposition. Second, C1$^*$ implies C1 by taking $D=C$ and applying weak union. Third, C1 and the fact that $Nd_G(D) = Nd_G(C)$ imply $D \ci_p Nd_G(D) \setminus Cc_G(Pa_G(C)) | Cc_G(Pa_G(C))$ by symmetry and decomposition, which together with C3$^*$ imply C1$^*$ by contraction. Finally, C2 and C2$^*$ are equivalent because $p(C | Pa_G(C)) = p(C | Cc_G(Pa_G(C)))$ by C1$^*$ and decomposition.
\end{proof}

Given $C \in Cc(G)$ and $D \subseteq C$, we define the marginal graph $G_C^D$ as the undirected graph over $D$ st $X - Y$ is in $G_C^D$ iff $X - Y$ is in $G_C$ or $X - V_1 - \ldots - V_n - Y$ is $G_C$ with $V_1, \ldots, V_n \notin D$.

\begin{lemma}\label{lem:factors}
Assume that $p$ is strictly positive and C1$^*$ holds. Then, C2$^*$ holds iff
\begin{equation}\label{eq:f2}
p(D | Pa_G(C)) = \prod_{K \in Cs(G^D_C)} \psi_D(K, Pa_G(K))
\end{equation}
for all $D \subseteq C$.
\end{lemma}

\begin{proof}
To prove the if part, it suffices to take $D = C$ and note that $G^D_C = G_C$. Then, C2$^*$ holds \cite[Proposition 3.8]{Lauritzen1996}. To prove the only if part, we adapt the proof of Theorem 3.9 by \citet{Lauritzen1996} to prove that $p(D | Pa_G(D))$ factorizes as indicated in Equation \ref{eq:f2}. This implies the desired result by C1$^*$ and decomposition. Specifically, choose arbitrary but fixed states $d^*$ and $pa_G(D)^*$ of $D$ and $Pa_G(D)$. Given $B \subseteq D$, let $\overline{b}^*$ and $\overline{pa_G(B)}^*$ denote the values of $D \setminus B$ and $Pa_G(D) \setminus Pa_G(B)$ consistent with $d^*$ and $pa_G(D)^*$. For all $B \subseteq D$, let
\[
H_D(b, pa_G(B)) = \log p(b, \overline{b}^* | pa_G(B), \overline{pa_G(B)}^*).
\]
Note that using the logarithm is warranted by the assumption of $p$ being strictly positive. For all $K \subseteq D$, let
\begin{equation}\label{eq:canonical}
\phi_D(k, pa_G(K)) = \sum_{B \subseteq K} (-1)^{|K \setminus B|} H_D(b, pa_G(B))
\end{equation}
where $b$ is consistent with $k$. Now, we can apply the M\"obius inversion \cite[Lemma A.2]{Lauritzen1996} to obtain
\[
\log p(d | pa_G(D)) = H_D(d, pa_G(D)) = \sum_{K \subseteq D} \phi_D(k, pa_G(K))
\]
where $k$ is consistent with $d$. Then, it only remains to prove that $\phi_D(k, pa_G(K))$ is zero whenever $K \notin Cs(G^D_C)$. Consider two nodes $S$ and $T$ of $K$ that are not adjacent in $G^D_C$. Then
\[
\phi_D(k, pa_G(K)) = \sum_{B \subseteq K \setminus S T} (-1)^{|(K \setminus S T) \setminus B|} \big[ H_D(b, pa_G(B)) - H_D(b s, pa_G(B S))
\]
\begin{equation}\label{eq:zero}
- H_D(b t, pa_G(B T)) + H_D(b s t, pa_G(B S T)) \big]
\end{equation}
where $b$, $bs$, $bt$ and $bst$ are consistent with $k$. Note that $S \ci_p T | (D \setminus S T) Pa_G(C)$ by C2$^*$, and $S \ci_p Pa_G(C) \setminus Pa_G(D) | (D \setminus S T) Pa_G(D)$ by C1$^*$, symmetry, decomposition and weak union. Then, $S \ci_p T | (D \setminus S T) Pa_G(D)$ by contraction and decomposition. This together with Equation 3.7 by \citet{Lauritzen1996} imply that
\[
H_D(b s t, pa_G(B S T)) - H_D(b s, pa_G(B S)) = \log \frac{p(bst, \overline{bst}^* | pa_G(B S T), \overline{pa_G(B S T)}^*)}{p(bs, \overline{bs}^* |  pa_G(B S), \overline{pa_G(B S)}^*)}
\]
\[
= \log \frac{p(s | b, \overline{bst}^*, pa_G(B S T), \overline{pa_G(B S T)}^*)p(bt, \overline{bst}^* | pa_G(B S T), \overline{pa_G(B S T)}^*)}{p(s | b, \overline{bst}^*, pa_G(B S), \overline{pa_G(B S)}^*)p(b, \overline{bs}^* | pa_G(B S), \overline{pa_G(B S)}^*)}.
\]
Moreover, note that $S \ci_p Pa_G(T) \setminus Pa_G(D \setminus T) | (D \setminus S T) Pa_G(D \setminus T)$ by C1$^*$, symmetry, decomposition and weak union. This implies that
\[
H_D(b s t, pa_G(B S T)) - H_D(b s, pa_G(B S))
\]
\[
= \log \frac{p(s | b, \overline{bst}^*, pa_G(B S), \overline{pa_G(B S T)}^*)p(bt, \overline{bst}^* | pa_G(B S T), \overline{pa_G(B S T)}^*)}{p(s | b, \overline{bst}^*, pa_G(B S), \overline{pa_G(B S T)}^*)p(b, \overline{bs}^* | pa_G(B S), \overline{pa_G(B S)}^*)}
\]
\[
= \log \frac{p(s^* | b, \overline{bst}^*, pa_G(B T), \overline{pa_G(B T)}^*)p(bt, \overline{bst}^* | pa_G(B S T), \overline{pa_G(B S T)}^*)}{p(s^* | b, \overline{bst}^*, pa_G(B T), \overline{pa_G(B T)}^*)p(b, \overline{bs}^* | pa_G(B S), \overline{pa_G(B S)}^*)}.
\]
Moreover, $S \ci_p Pa_G(T) \setminus Pa_G(D \setminus T) | (D \setminus S T) Pa_G(D \setminus T)$ also implies that
\[
H_D(b s t, pa_G(B S T)) - H_D(b s, pa_G(B S))
\]
\[
= \log \frac{p(s^* | b, \overline{bst}^*, pa_G(B T), \overline{pa_G(B T)}^*)p(bt, \overline{bst}^* | pa_G(B S T), \overline{pa_G(B S T)}^*)}{p(s^* | b, \overline{bst}^*, pa_G(B), \overline{pa_G(B)}^*)p(b, \overline{bs}^* | pa_G(B S), \overline{pa_G(B S)}^*)}.
\]
Finally, note that $D \setminus S \ci_p Pa_G(S) \setminus Pa_G(D \setminus S) | Pa_G(D \setminus S)$ by C1$^*$ and decomposition. This implies that
\[
H_D(b s t, pa_G(B S T)) - H_D(b s, pa_G(B S))
\]
\[
= \log \frac{p(s^* | b, \overline{bst}^*, pa_G(B T), \overline{pa_G(B T)}^*)p(bt, \overline{bst}^* | pa_G(B T), \overline{pa_G(B T)}^*)}{p(s^* | b, \overline{bst}^*, pa_G(B), \overline{pa_G(B)}^*)p(b, \overline{bs}^* | pa_G(B), \overline{pa_G(B)}^*)}
\]
\[
= \log \frac{p(bt, \overline{bt}^* | pa_G(B T), \overline{pa_G(B T)}^*)}{p(b, \overline{b}^* |  pa_G(B), \overline{pa_G(B)}^*)} = H_D(b t, pa_G(B T)) - H_D(b, pa_G(B)).
\]
Thus, all the terms in the square brackets in Equation \ref{eq:zero} add to zero, which implies that the entire sum is zero.
\end{proof}

It is customary to think of the factors $\psi_D(K, Pa_G(K))$ in Equation \ref{eq:f2} as arbitrary non-negative functions, whose product needs to be normalized to result in a probability distribution. Note however that Equation \ref{eq:f2} does not include any normalization constant. The reason is that the so called canonical parameterization in Equation \ref{eq:canonical} permits us to write any probability distribution as a product of factors that does not need subsequent normalization. One might think that this must be an advantage for parameter estimation and inference. However, the truth is that the cost of computing the normalization constant has been replaced by the cost of having to manipulate a larger number of factors in Equation \ref{eq:f2}. To see it, note that the size of $Cs(G^D_C)$ is exponential in the size of the largest clique in $G^D_C$.

A necessary and sufficient factorization follows.

\begin{theorem}\label{the:equivalence0}
Let $p$ be a strictly positive probability distribution. Then, $p$ is Markovian wrt an AMP CG $G$ iff
\begin{equation}\label{eq:e0}
p(V) = \prod_{C \in Cc(G)} p(C | Pa_G(C))
\end{equation}
with
\begin{equation}\label{eq:e02}
p(D | Pa_G(C)) = \prod_{K \in Cs(G^D_C)} \psi_D(K, Pa_G(K))
\end{equation}
for all $D \subseteq C$.
\end{theorem}

\begin{proof}
The only if part holds because C1$^*$ and decomposition imply Equation \ref{eq:e0}, and Lemma \ref{lem:factors} implies Equation \ref{eq:e02}. To prove the if part, we prove that $p$ satisfies C1$^*$ and C2$^*$. Note that $Nd_G(C) = Nd_G(D)$. This together with Equations \ref{eq:e0} and \ref{eq:e02} imply that
\[
p(D, Nd_G(D)) = p(D, Nd_G(C)) = \bigg( \prod_{U \in Cc(G): U \subseteq Nd_G(C)} p(U | Pa_G(U)) \bigg) p(D | Pa_G(C))
\]
\[
= g(Nd_G(D)) h(D, Pa_G(D))
\]
and thus C1$^*$ holds \cite[Equation 3.6]{Lauritzen1996}. Finally, C2$^*$ holds by Equation \ref{eq:e02} and Lemma \ref{lem:factors}.
\end{proof}

A more convenient necessary and sufficient factorization follows.

\begin{theorem}\label{the:equivalence}
Let $p$ be a strictly positive probability distribution. Then, $p$ is Markovian wrt an AMP CG $G$ iff
\begin{equation}\label{eq:e}
p(V) = \prod_{C \in Cc(G)} p(C | Pa_G(C))
\end{equation}
with
\begin{equation}\label{eq:e2}
p(C | Pa_G(C)) = \prod_{K \in Cs(G_C)} \psi_C(K, Pa_G(K))
\end{equation}
and
\begin{equation}\label{eq:e3}
p(D | Pa_G(C)) = p(D | Pa_G(D))
\end{equation}
for all $D \subseteq C$.
\end{theorem}

\begin{proof}
The only if part holds because C1$^*$ and decomposition imply Equations \ref{eq:e} and \ref{eq:e3}, and Lemma \ref{lem:factors} implies Equation \ref{eq:e2}. To prove the if part, we prove that $p$ satisfies C1$^*$ and C2$^*$. Note that $Nd_G(C) = Nd_G(D)$. This together with Equations \ref{eq:e} and \ref{eq:e3} imply that
\[
p(D, Nd_G(D)) = p(D, Nd_G(C)) = \bigg( \prod_{U \in Cc(G): U \subseteq Nd_G(C)} p(U | Pa_G(U)) \bigg) p(D | Pa_G(C))
\]
\[
= \bigg( \prod_{U \in Cc(G): U \subseteq Nd_G(C)} p(U | Pa_G(U)) \bigg) p(D | Pa_G(D)) = g(Nd_G(D)) h(D, Pa_G(D))
\]
and thus C1$^*$ holds \cite[Equation 3.6]{Lauritzen1996}. Finally, C2$^*$ holds by Equation \ref{eq:e2} \cite[Proposition 3.8]{Lauritzen1996}.
\end{proof}

A necessary factorization that is more convenient for inference and parameter learning follows.

\begin{corollary}\label{cor:equivalence}
Let $p$ be a strictly positive probability distribution. If $p$ is Markovian wrt an AMP CG $G$, then
\begin{equation}\label{eq:ec}
p(V) = \prod_{C \in Cc(G)} p(C | Pa_G(C))
\end{equation}
with
\begin{equation}\label{eq:ec2}
p(C | Pa_G(C)) = \prod_{K \in Cs(G_C)} \psi_C(K, Pa_G(K)).
\end{equation}
\end{corollary}

\section{Parameter Learning}

Given some data, we can efficiently obtain the maximum likelihood estimates of the factors in Equation \ref{eq:ec2} by adapting the iterative proportional fitting procedure (IPFP) for MRFs \cite[Section 19.5.7]{Murphy2012} as follows:
\begin{center}
\begin{tabular}{rl}
1 & For each $C \in Cc(G)$\\
2 & \hspace{0.25cm} Set $p^0(C | Pa_G(C))$ to the uniform distribution\\
3 & \hspace{0.25cm} Compute $\phi_C(K, Pa_G(K))$ for all $K \in Cs(G_C)$ as shown in Equation \ref{eq:canonical}\\
4 & \hspace{0.25cm} Set $\psi_C(K, Pa_G(K)) = \exp \phi_C(K, Pa_G(K))$ for all $K \in Cs(G_C)$\\
5 & \hspace{0.25cm} Repeat until convergence\\
6 & \hspace{0.5cm} Set $\psi_C(K, Pa_G(K)) = \psi_C(K, Pa_G(K)) \frac{p_e(K | Pa_G(K))}{p(K | Pa_G(K))}$ for all $K \in Cs(G_C)$\\
\end{tabular}
\end{center}
where $p_e$ is the empirical probability distribution over $V$ obtained from the given data, and $p$ is the probability distribution over $V$ due to the current estimates. Note that computing $p(K | Pa_G(K))$ requires inference. The multiplication and division in line 6 are elementwise. Existing gradient ascend methods for MRFs can be adapted similarly.

We justify the algorithm above by adapting some existing results for MRFs. We temporally drop the assumption that the product of factors in Equation \ref{eq:ec2} is normalized, and replace it with
\begin{equation}\label{eq:e2-b}
p(C | Pa_G(C)) = \frac{1}{Z_C(Pa_G(C))} \prod_{K \in Cs(G_C)} \psi_C(K, Pa_G(K))
\end{equation}
where
\[
Z_C(Pa_G(C)) = \sum_c \prod_{K \in Cs(G_C)} \psi_C(k, Pa_G(K))
\]
where $k$ is consistent with $c$. Let $\psi$ denote all the factors due to Equations \ref{eq:ec} and \ref{eq:e2-b}. Then, the log-likelihood function is
\[
l(\psi) = \sum_{C \in Cc(G)} \bigg( \sum_{K \in Cs(G_C)} \sum_k \sum_{pa_G(K)} n(k, pa_G(K)) \log \psi_C(k, pa_G(K)) - n(pa_G(C)) \log Z_C(pa_G(C)) \bigg)
\]
where $n(k, pa_G(K))$ is the number of instances in the data where $K$ and $Pa_G(K)$ take values $k$ and $pa_G(K)$ simultaneously. Similarly for $n(pa_G(C))$. Dividing both sides by the number of instances in the data, $n$, we have that
\[
l(\psi) / n = \sum_{C \in Cc(G)} \bigg( \sum_{K \in Cs(G_C)} \sum_k \sum_{pa_G(K)} p_e(k, pa_G(K)) \log \psi_C(k, pa_G(K)) - p_e(pa_G(C)) \log Z_C(pa_G(C)) \bigg).
\]
Let $U \in Cc(G)$ and $Q \in Cs(G_U)$. The gradient of $l(\psi) / n$ wrt $\psi_U(q, pa_G(Q))$ is
\[
\frac{\partial l(\psi) / n}{\partial \psi_U(q, pa_G(Q))} = \frac{p_e(q, pa_G(Q))}{\psi_U(q, pa_G(Q))} - \frac{p_e(pa_G(U))}{Z_U(pa_G(U))} \frac{\partial Z_U(pa_G(U))}{\partial \psi_U(q, pa_G(Q))}.
\]
Let $W = U \setminus Q$. Then
\[
\frac{\partial Z_U(pa_G(U))}{\partial \psi_U(q, pa_G(Q))} = \sum_w \prod_{K \in Cs(G_U) \setminus Q} \psi_U(k, \overline{k}, pa_G(K))
\]
\[
= \frac{Z_U(pa_G(U))}{\psi_U(q, pa_G(Q))} \sum_w \prod_{K \in Cs(G_U) \setminus Q} \psi_U(k, \overline{k}, pa_G(K)) \frac{\psi_U(q, pa_G(Q))}{Z_U(pa_G(U))} = \frac{Z_U(pa_G(U))}{\psi_U(q, pa_G(Q))} p(q | pa_G(U))
\]
where $\overline{k}$ denotes the elements of $q$ corresponding to the elements of $K \cap Q$, and the last equality follows from Equation \ref{eq:e2-b}. Note also that $p(q | pa_G(Q)) = p(q | pa_G(U))$ by C1$^*$ and decomposition. Putting together the results above, we have that
\[
\frac{\partial l(\psi) / n}{\partial \psi_U(q, pa_G(Q))} = \frac{p_e(q, pa_G(Q))}{\psi_U(q, pa_G(Q))} - \frac{p_e(pa_G(Q)) p(q | pa_G(Q))}{\psi_U(q, pa_G(Q))}.
\]
Since the maximum likelihood estimates are obtained when the gradient is 0 for all the entries of all the factors, we have that the maximum likelihood estimates are obtained when
\[
\psi_C(k, pa_G(K)) = \psi_C(k, pa_G(K)) \frac{p_e(k | pa_G(K))}{p(k | pa_G(K))}
\]
for all $C \in Cc(G)$ and $K \in Cs(G_C)$. This justifies the updating step in line 6 of the IPFP.

Let the factor updated in the current iteration have the superscript $t+1$, whereas the rest of the factors have the superscript $t$. Next, we show that if $Z_C^0(Pa_G(C)) = 1$ then $Z_C^{t+1}(Pa_G(C)) = 1$. This implies that Equations \ref{eq:e2} and \ref{eq:e2-b} are equivalent because $Z_C^0(Pa_G(C)) = 1$ by line 3 and, thus, our assumption of Equation \ref{eq:e2-b} is innocuous. To see it, let $U \in Cc(G)$, $Q \in Cs(G_U)$ and $W = C \setminus Q$. Then
\[
p^{t+1} (Q | Pa_G(U)) = \sum_w p^{t+1} (w | Pa_G(U))
\]
\[
= \sum_w \psi_U^{t+1}(Q, Pa_G(Q)) \frac{1}{Z_U^{t+1}(Pa_G(U))} \prod_{K \in Cs(G_U) \setminus Q} \psi_U^t(k, Pa_G(K))
\]
\[
= \sum_w \psi_U^t(Q, Pa_G(Q)) \frac{p_e(Q | Pa_G(Q))}{p^t(Q | Pa_G(Q))} \frac{1}{Z_U^{t+1}(Pa_G(U))} \prod_{K \in Cs(G_U) \setminus Q} \psi_U^t(k, Pa_G(K))
\]
\[
= \frac{p_e(Q | Pa_G(Q))}{p^t(Q | Pa_G(Q))} \frac{Z_U^t(Pa_G(U))}{Z_U^{t+1}(Pa_G(U))} \sum_w p^t (w | Pa_G(U)) = \frac{p_e(Q | Pa_G(Q))}{p^t(Q | Pa_G(Q))} \frac{Z_U^t(Pa_G(U))}{Z_U^{t+1}(Pa_G(U))} p^t (Q | Pa_G(U))
\]
\[
= p_e(Q | Pa_G(Q)) \frac{Z_U^t(Pa_G(U))}{Z_U^{t+1}(Pa_G(U))}
\]
since $p^t(Q | Pa_G(Q)) = p(Q | Pa_G(U))$ by C1$^*$ and decomposition. Summing both sides over $q$ implies that the IPFP preserves the normalization constant across iterations.

\section{Discussion}

Given a probability distribution $p$ that is Markovian wrt an AMP CG $G$, we have described in Equations \ref{eq:e}-\ref{eq:e3} necessary and sufficient conditions for $p$ to factorize wrt $G$. This note extends the original manuscript, where $p$ is shown to factorize as
\[
p(V) = \prod_{C \in Cc(G)} \prod_{K \in Cs(G_C)} \psi_C(K, Pa_G(K)).
\]
To see that the condition above is necessary but not sufficient, consider the AMP CGs $A \ra B - C$ and $A \ra B - C \la A$, and note that both imply the same factorization, namely $p(A,B,C) = \psi_A(A) \psi_{BC}(A, B, C)$. So, if $p$ encodes no independence then it factorizes according to both CGs although it is Markovian wrt only the second of them. In any case, the factorization above is enough to perform efficiently inference and parameter estimation, as shown in the original manuscript.

Unfortunately, finding the maximum likelihood estimates of the factors in the new factorization is difficult and, thus, we have decided to enforce only Equations \ref{eq:e} and \ref{eq:e2} in the estimation process so that it can be performed efficiently via the IPFP. The so fitted factorization is enough to perform inference efficiently following the same procedure as in the original manuscript.

Our work is related to that by \citet{Drton2008}, where the author proposes necessary and sufficient conditions for $p$ to factorize wrt $G$ when $G$ is a MVR CG. His factorization resembles ours in that it includes constraints similar to those in Equation \ref{eq:e2}, which make maximum likelihood estimation hard. To overcome this problem, the author develops a so called iterative conditional fitting procedure (ICFP) that, at each iteration, solves a convex optimization problem under the mentioned constraints. We plan to study whether it is possible to adapt the ICFP to our problem, given the similarity between the constraints in both factorizations. \citet{Drton2008} also makes the interesting observation that the runtime of the ICFP can be shortened by replacing $G$ with a Markov equivalent CG with smaller connectivity components. It would be interesting to see whether this also applies to our IPFP. A result that would be helpful in that investigation is that by \citet[Theorem 4]{SonntagandPenna2015}, which shows how to obtain a Markov equivalent CG with the fewest undirected edges.

Two other works that are related to ours are those by \citet{Abbeeletal.2006} and \citet{Royetal.2009}. Unlike our work, these works do not characterize when the Markovian and factorization properties are equivalent. Instead, they develop closed form expressions for estimating the factors in the factorization of $p$ wrt $G$ when $G$ is a factor graph. Since factor graphs subsume AMP CGs, we can adapt their closed form estimates to our problem. Specifically, let $C \in Cc(G)$ and $K \in Cs(G_C)$. Also, let $Mb_G(K)$ denote the minimal subset of $C Pa_G(C)$ st $K \ci_G C Pa_G(C) \setminus K Mb_G(K) | Mb_G(K)$. It is easy to see that $Mb_G(K) = Ne_G(K) Pa_G(K) Pa_G(Ne_G(K))$. Now, choose an arbitrary but fixed state $v^*$ of $V$. Then, we can use Proposition 4 by \cite{Abbeeletal.2006} to rewrite the factorization in Corollary \ref{cor:equivalence} as follows (details omitted):
\begin{equation}\label{eq:a}
p(v) = p(v^*) \prod_{C \in Cc(G)} \prod_{K \in Cs(G_C) : K \neq \emptyset} \psi(k)
\end{equation}
where $k$ is consistent with $v$, and
\begin{equation}\label{eq:a2}
\psi(k) = \exp \bigg( \prod_{B \subseteq K} (-1)^{|K \setminus B|} \log p(b, \overline{b}^* | mb_G(K)^*) \bigg)
\end{equation}
where $\overline{b}^*$ and $mb_G(K)^*$ denote the values of $K \setminus B$ and $Mb(K)$ consistent with $v^*$. In order to estimate the factors above, the authors propose replacing $p$ with the empirical probability distribution $p_e$. Unfortunately, this may produce an unreliable estimate for $p(v^*)$ unless the data available is abundant. Similarly for $p(b, \overline{b}^* | mb_G(K)^*)$ because $K$ and $Mb_G(K)$ may be large. Note also that the estimate of $p(b, \overline{b}^* | mb_G(K)^*)$ is based only on the instances of the data that are consistent with $\overline{b}^*$ and $mb_G(K)^*$ simultaneously. This means that the data available is not used efficiently. All this leads the authors to acknowledge that their closed form estimates, as described, are probably impractical \cite[p. 1764]{Abbeeletal.2006}. \citet{Royetal.2009} improve the method above by simplifying Equation \ref{eq:a2}. Although the improvement alleviates the drawbacks mentioned, it does not eliminate them completely (e.g. Equation \ref{eq:a} stays the same and, thus, the problem of estimating $p(v^*)$ remains). Unfortunately, no experimental results are reported in either of the works cited. It would be interesting to compare them with our IPFP.